\documentclass[10pt,twocolumn,letterpaper]{article}

\usepackage{cvpr}
\usepackage{times}
\usepackage{epsfig}
\usepackage{graphicx}
\usepackage{amsmath}
\usepackage{amssymb}
\usepackage{listings}
\usepackage{enumerate}
\usepackage{gensymb}
\usepackage[caption = false]{subfig}
\usepackage{float}
\usepackage{multirow}

\usepackage[pagebackref=true,breaklinks=true,letterpaper=true,colorlinks,bookmarks=false]{hyperref}
\author{Noureldin Hendy , Cooper Sloan, Feng Tian \\ Pengfei Duan, Nick Charchut, Yuesong Xie, Chuang Wang, James Philbin\\
Zoox\\
{\tt\small \{nhendy,csloan,ftian,pduan,ncharchut,yxie,mrwang,james\}@zoox.com}
}
\cvprfinalcopy 

\ifcvprfinal\fi
\begin{document}

\title{FISHING Net: Future Inference of Semantic Heatmaps In Grids}

\maketitle

\begin{abstract}
 For autonomous robots to navigate a complex environment, it is crucial to understand the surrounding scene both geometrically and semantically. Modern autonomous robots employ multiple sets of sensors, including lidars, radars, and cameras. Managing the different reference frames and characteristics of the sensors, and merging their observations into a single  representation complicates perception. Choosing a single unified representation for all sensors simplifies the task of perception and fusion. In this work, we present an end-to-end pipeline that performs semantic segmentation and short term prediction using a top-down representation. Our approach consists of an ensemble of neural networks which take in sensor data from different sensor modalities and transform them into a single common top-down semantic grid representation. We find this representation favorable as it is agnostic to sensor-specific reference frames and captures both the semantic and geometric information for the surrounding scene. Because the modalities share a single output representation, they can be easily aggregated to produce a fused output. In this work we predict short-term semantic grids but the framework can be extended to other tasks.  This approach offers a simple, extensible, end-to-end approach for multi-modal perception and prediction. 

\end{abstract}

\section{Introduction} %

A key task for autonomous robots is \emph{perception}, which involves building a representation that captures the geometry and semantics of the surrounding scene. It is crucial that such a representation is accurate enough to be used by an autonomous robot to plan and make decisions to reach a goal. Autonomous robots commonly use multiple complimentary sensors such as lidars, radars, and cameras. These sensors vary significantly in their working principles, output representations, and signal qualities. Thus it is desirable to unify the information extracted from all sensors into one common representation. A top-down representation is natural for geometric scene understanding. Top-down representations compactly aggregates the geometric and semantic configurations of different agents in a scene. 
We leverage this representation to serve as a common output interface between sensor modalities. The benefits of a shared top-down representation across modalities are threefold. First, it is an interpretable representation that better facilitates debugging and reasoning about inherent failure modes of each modality. Second, it is independent of any particular sensor’s characteristics and so is easily extensible for adding new modalities. Finally, it simplifies the task of late fusion by sharing a spatial representation in a succinct manner.

\begin{figure}[t]
\begin{center}
\includegraphics[width=1.0\linewidth]{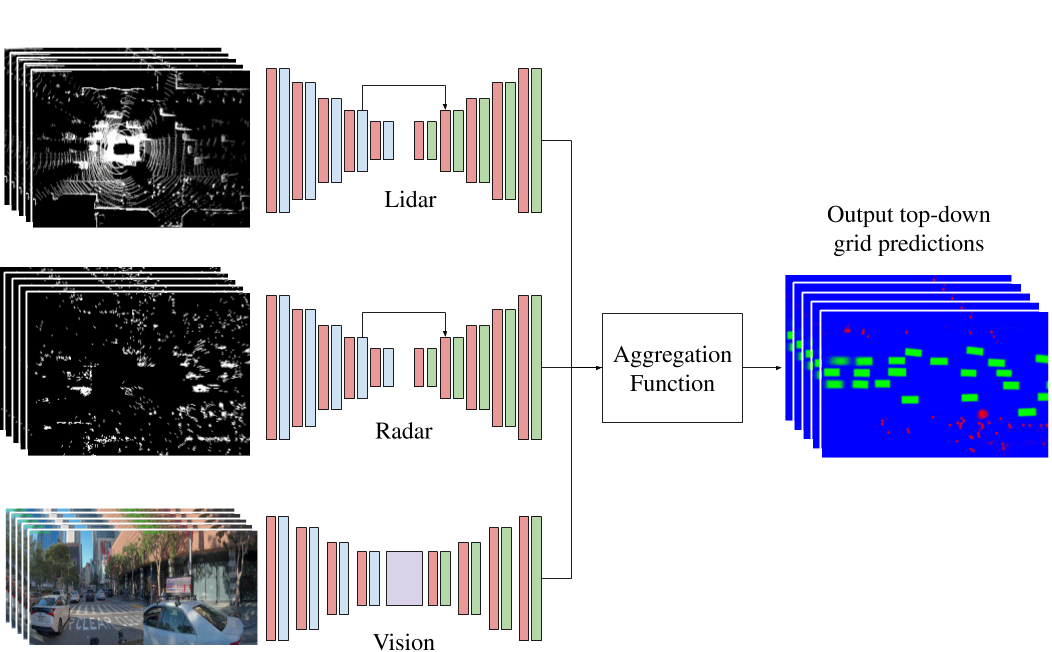}
\end{center}
\caption{FISHING Net Architecture: multiple neural networks, one for each sensor modality (lidar, radar and camera) take in a sequence of input sensor data from $t=-2,-1.5.-1,-0.5,0s$ and output a sequence of shared top-down semantic grids at $t=0,0.5,1,1.5,2s$ representing 3 object classes (Vulnerable Road Users (VRU), vehicles and background). The sequences are then fused using an aggregation function to output a
fused sequence of semantic grids.}
\label{fishingnet}
\end{figure}

In addition to perceiving the current state of the world, autonomous robots must be able to estimate future states in order to make effective decisions toward their destinations. This key task is commonly referred to as \emph{behavior prediction}. Traditional pipelines typically employ separate tracking and trajectory prediction modules to achieve this task \cite{PerceptionPlanningControl_2017}, consequently increasing system complexity and latency. In this work, we show that using a shared top-down representation as an output from different sensor modalities, it is possible to not only \emph{perceive} the current state of the environment, but also \emph{predict} future states within a short time horizon in an end-to-end fashion. This end-to-end perception and prediction formulation reduces computation cost, system complexity and overall latency compared to a modular pipeline.

In this work we present a novel end-to-end framework that predicts the top-down view of the current scene ($t_0$) as well as multiple timesteps into the future. The pipeline consists of a convolutional neural network for each of three sensor modalities: lidar, radar, and camera. Each sensor modality predicts a sequence of top-down semantic grids, then these outputs are fused to produce a single output grid. We explore fusing using two different aggregation mechanisms. We evaluate our pipeline using only cameras and lidar modalities on NuScenes dataset \cite{caesar2019nuscenes} and Lyft dataset \cite{lyft2019}. We also evaluate on a purpose-built dataset collected internally and use all three sensor modalities. 

Our main contributions are the following:

\begin{enumerate}[i]
\setlength{\itemsep}{0pt}
\setlength{\parskip}{0pt}
\setlength{\parsep}{0pt}
 \item We propose a novel multi-modal architecture to perform end-to-end perception and prediction.
  \item We propose a novel application of using a top-down semantic grid as a common output interface to facilitate late sensor fusion. 
 \item We propose a novel architecture for short-term prediction of top-down semantic grids using Radar data. 
\item We show that this architecture is extensible to 2D and 3D sensors in varied reference frames.
\end{enumerate}

\section{Related Work}

\textbf{Semantic Segmentation} 
In semantic segmentation a neural network is trained to predict a mask which assigns a semantic label to each pixel in the input image. There has been a considerable body of literature dedicated to predicting a semantic mask on the perspective view of the image including FCN \cite{FCNPaper}, SegNet \cite{SegNetPaper}, U-Net \cite{UNetPaper}, and PSPNet \cite{PSPNetPaper}. In this work however, we predict a semantic mask in the top-down view of the scene, rather than in pixel space. There has been recent research attempting to achieve similar tasks \cite{MonocularED} \cite{roddick2018orthographic} \cite{Palazzi2017LearningTM} . These approaches use a single camera, or transform a detected bounding box to a top-down view and only predict the current timestep. Closest to our work is the work done by Pan et al. \cite{CrossView} where they introduce a framework called View Parsing Network (VPN) to achieve a task they call Cross View segmentation. The input is an image from multiple cameras situated at different angles and the output is a top-down view of the scene. We employ a similar approach and extend it to predict multiple timesteps into the future specifically for cameras. Erkent et al. \cite{End2EndSemanticGrid} predict a semantic grid from camera images but perform an early fusion of occupancy grid computed using lidar data and evaluate in a single camera setting. Hoyer et al.\cite{Hoyer2019ShortTermPA} predict a top-down semantic grid representation and use images from cameras situated at different angles. However, their pipeline first runs semantic segmentation on the images and then uses stereo-depth in order to map each pixel’s semantic label to a top-down grid, whereas we operate directly on raw camera images which reduces pipeline complexity. Prophet et al.\cite{RadarSemseg} present a framework for generated top-down semantic grids using radar data. Prophet et al. focus on inferring the current semantic grid, while our work goes further to predict the future positions of surrounding agents.

\textbf{Short-term Prediction} There is recent research performing short-term prediction of grids; \cite{MultistepOGM}, \cite{LongTermPrediction}, \cite{DeepTracking} propose using a recurrent neural network that takes as input a sequence of occupancy grids generated using lidar data and predicts occupancy grids into the future. We show that we can predict more than just occupied and non-occupied classes. Our semantic grid consists of vulnerable road users (pedestrians, bicyclists, motorists, etc), vehicles, and background classes. Additionally, we further apply the same formulation to other modalities such as cameras and radar. Close to our work is that done by \cite{Hoyer2019ShortTermPA} which proposes a pipeline that predicts a semantic grid at a future timestep. We distinguish ourselves in that we not only predict the current timestep but multiple future timesteps.

\section{FISHING Net}

\subsection{Problem Formulation}
\begin{figure}[h]
   \begin{center}
   \includegraphics[width=0.8\linewidth]{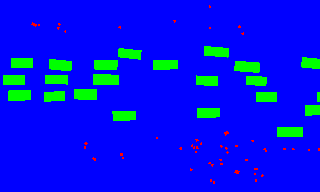}
   \caption{Top-down view ground truth semantic grid. (Vehicles: Green; VRU: Red; Background: Blue).}.
   \label{fig:ground_truth}
   \end{center}
   \end{figure}

\begin{figure}[h]
   \begin{center}
   \includegraphics[width=0.8\linewidth]{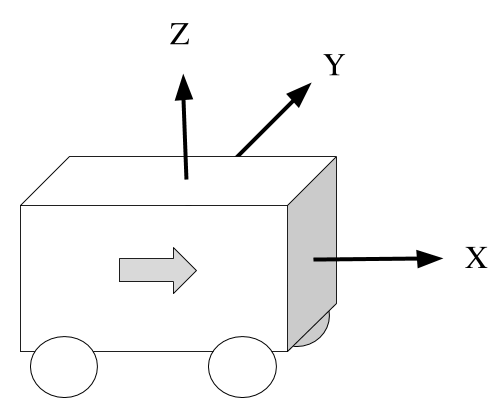}
   \caption{Top-Down Reference Frame}.
   \label{fig:reference_frame}
   \end{center}
   \end{figure}

We formulate the task as follows: 
Given a set of $k$ sensor modalities with sensory data from the current timestep $t_0$ and $p$ past timesteps $t_{-p}$, ..., $t_{-1}$, infer top-down semantic grids at the current timestep $t_0$ and $f$ future timesteps $t_1$, ..., $t_{f}$
Three object classes are represented in our semantic grids: 1) vulnerable road users (VRU) including pedestrians, bicyclists and motorists, 2) vehicles, and 3) background. An example of the representation for a single timestep is shown in Figure \ref{fig:ground_truth}.  
To compensate for ego vehicle motion we pick a fixed reference frame centered on the ego position at $t_0$. The $x$ dimension represents the forward direction of travel, the $y$ axis at a right angle pointing left out of the robot, and the $z$ axis is the vertical height dimension, show in Figure \ref{fig:reference_frame}.


We use a network for each of the $k$ modalities to do short term prediction of semantic grids into the future.
Each network consumes multiple timesteps of sensory data and outputs multiple future timesteps of top-down semantic grids. The semantic grids are then aggregated across modalities to produce a fused grid. 
This output grid provides a top-down view of the dynamic agents surrounding the ego vehicle.
The overall architecture using lidar, radar and cameras as the input modalities is described in Figure \ref{fishingnet}.

\subsection{Lidar Input}

Our input lidar representation consists of 8 top-down grid features, matching the output dimensions, resolution, and reference frame. The features are generated by projecting lidar points into the top-down grid reference frame and computing various functions on the set of points in each grid cell. 

These features consist of: 1) Binary lidar occupancy (1 if any lidar point is present a given grid cell, 0 otherwise). 2) Lidar density (Log normalized density of all lidar points present in a grid cell). 3) Max z (Largest height value for lidar points in a given grid cell). 4) Max z sliced (Largest z value for each grid cell over 5 linear slices eg. 0-0.5m, ..., 2.0-2.5m).
These features are stacked along the channel dimension for the purposes of training.
For a single input frame the dimensions of the Lidar features would be $\mathbb{R}^{W \times H \times 8}$. For a sequence of input frames we'd end up with $\mathbb{R}^{W \times H \times 8 \times (p+1)}$ where $p+1$ is the number of input frames. 
{
\begin{figure}
\subfloat[]{\includegraphics[width = 0.45\linewidth]{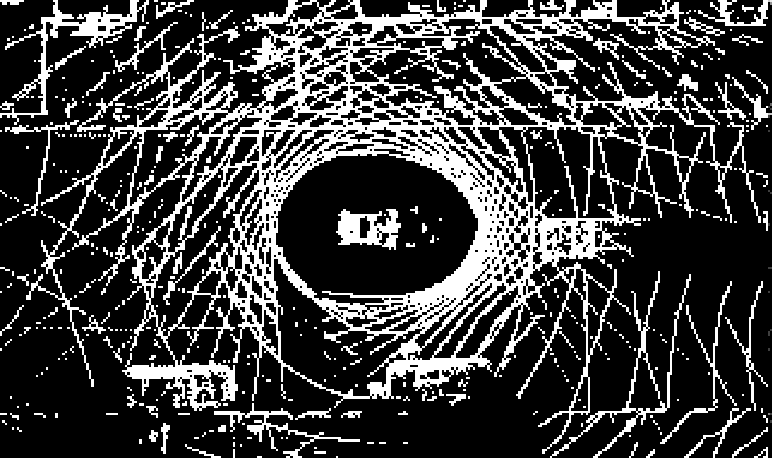}}
    \hspace{0.1\linewidth}
\subfloat[]{\includegraphics[width = 0.45\linewidth]{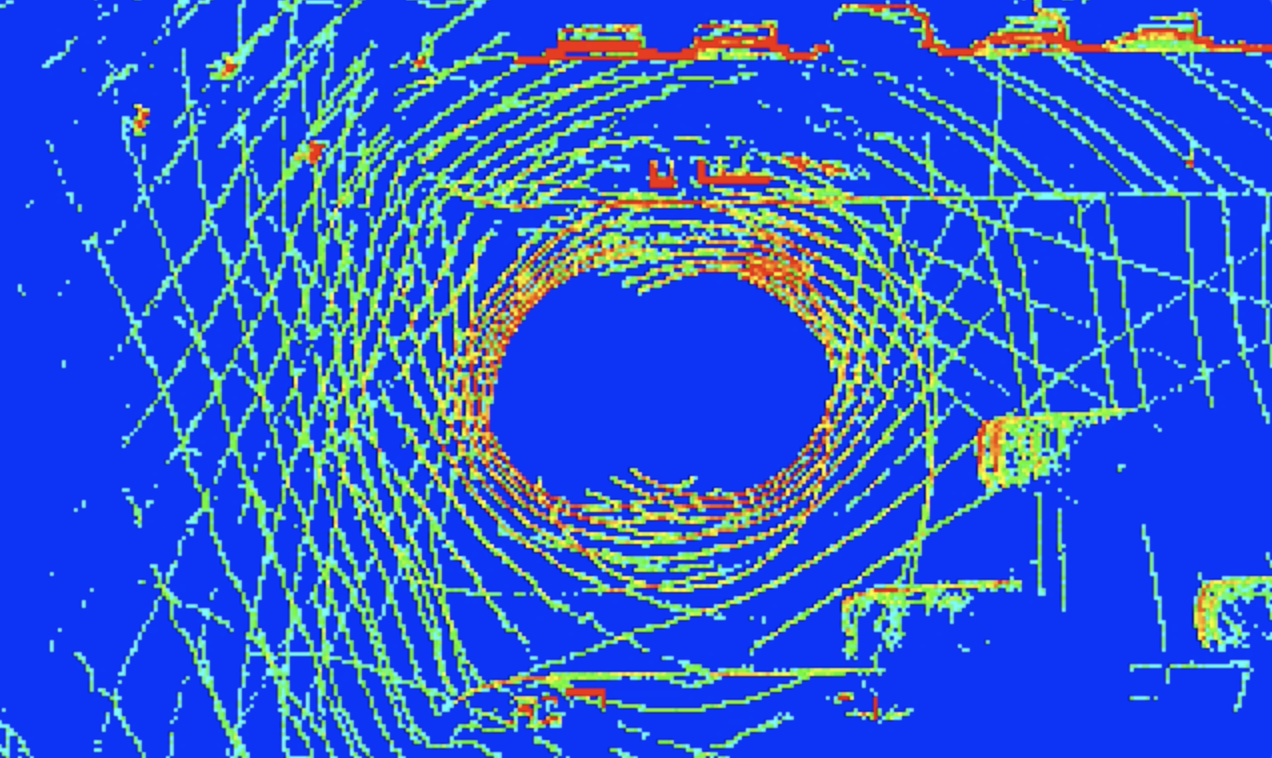}}\\
\subfloat[]{\includegraphics[width = 0.45\linewidth]{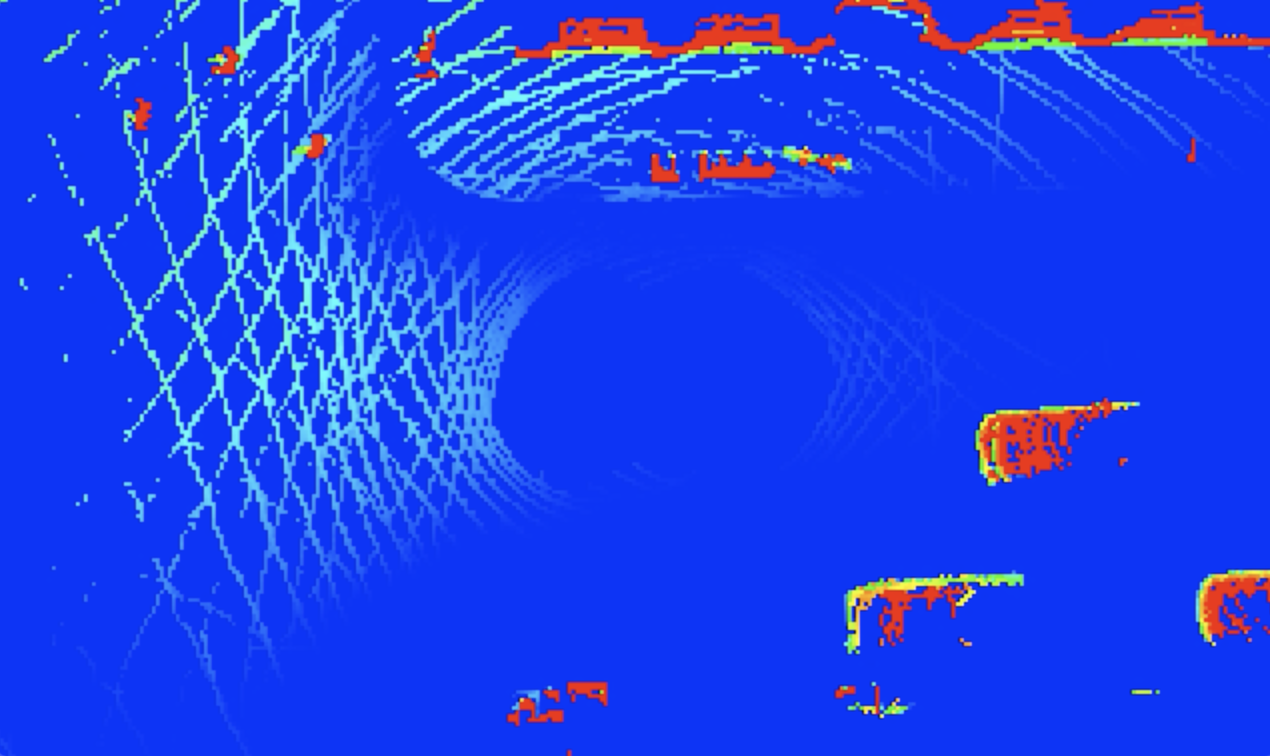}}
    \hspace{0.1\linewidth}
\subfloat[]{\includegraphics[width = 0.45\linewidth]{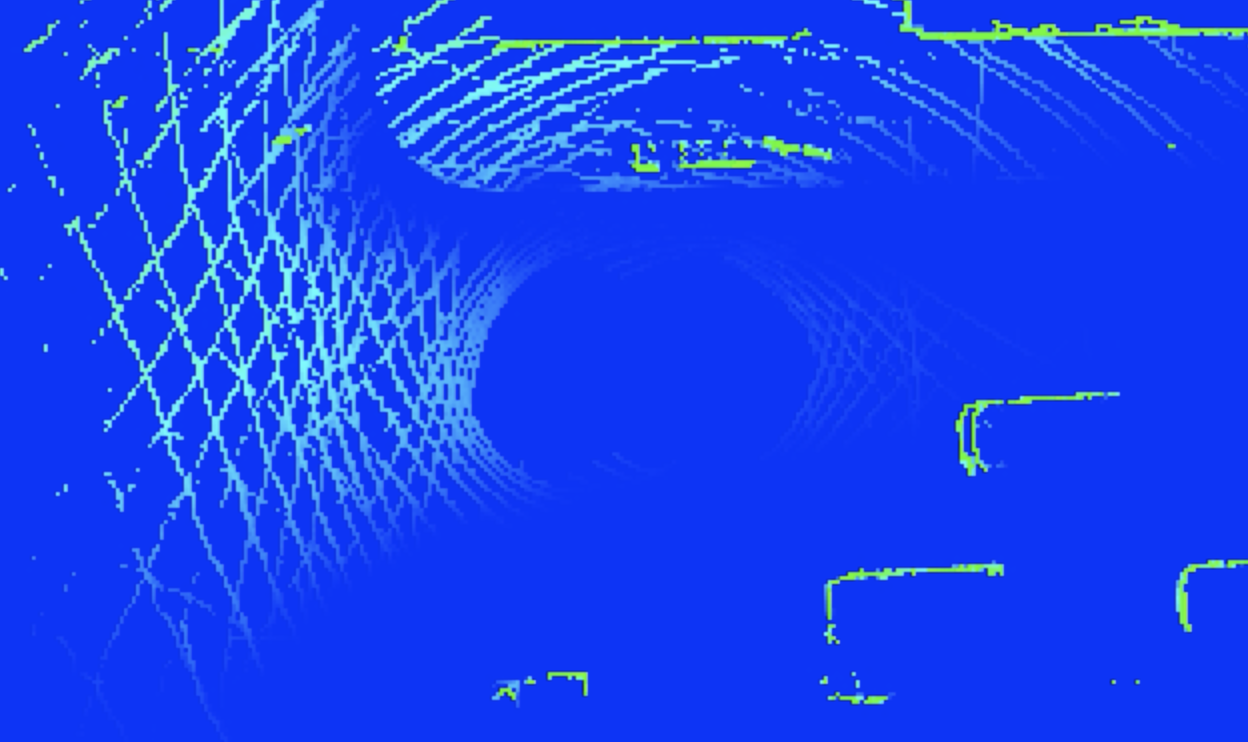}}
    \caption{Subset of lidar features extracted at current timestep. (a) Lidar occupancy, white is occupied and black is available. 
    (b) Lidar density. (c) Maximum $Z$ measurement. (d) Maximum $Z$ measurement for 1st height slice}
\label{fig:lidar features}
\end{figure}

\begin{figure}[t]
\begin{center}
\includegraphics[width=0.9\linewidth]{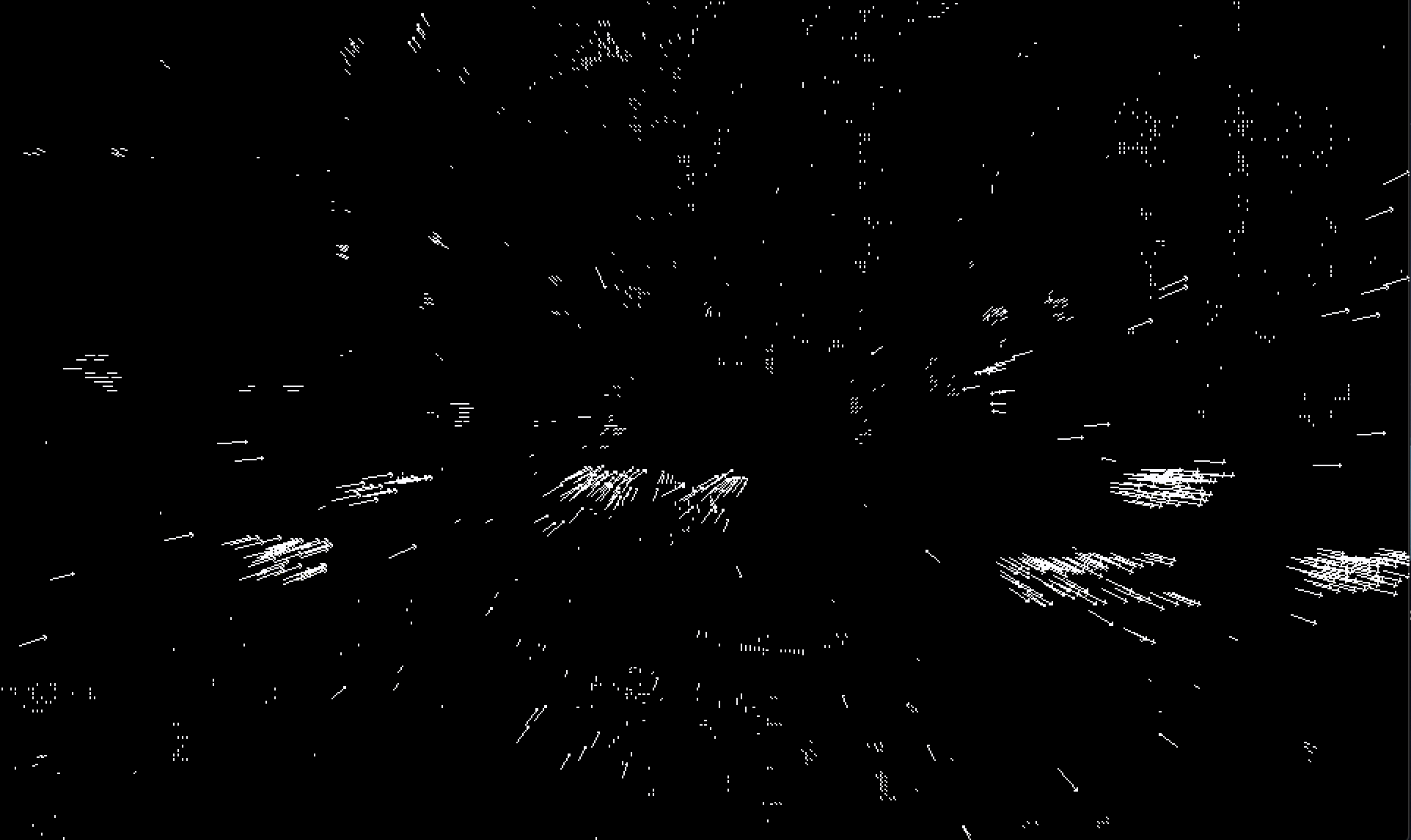}
\end{center}
\caption{Quiver plot for a single frame velocity features from radars}
\label{radar_input}
\end{figure}

\begin{figure}[t]
\begin{center}
\includegraphics[width=0.9\linewidth]{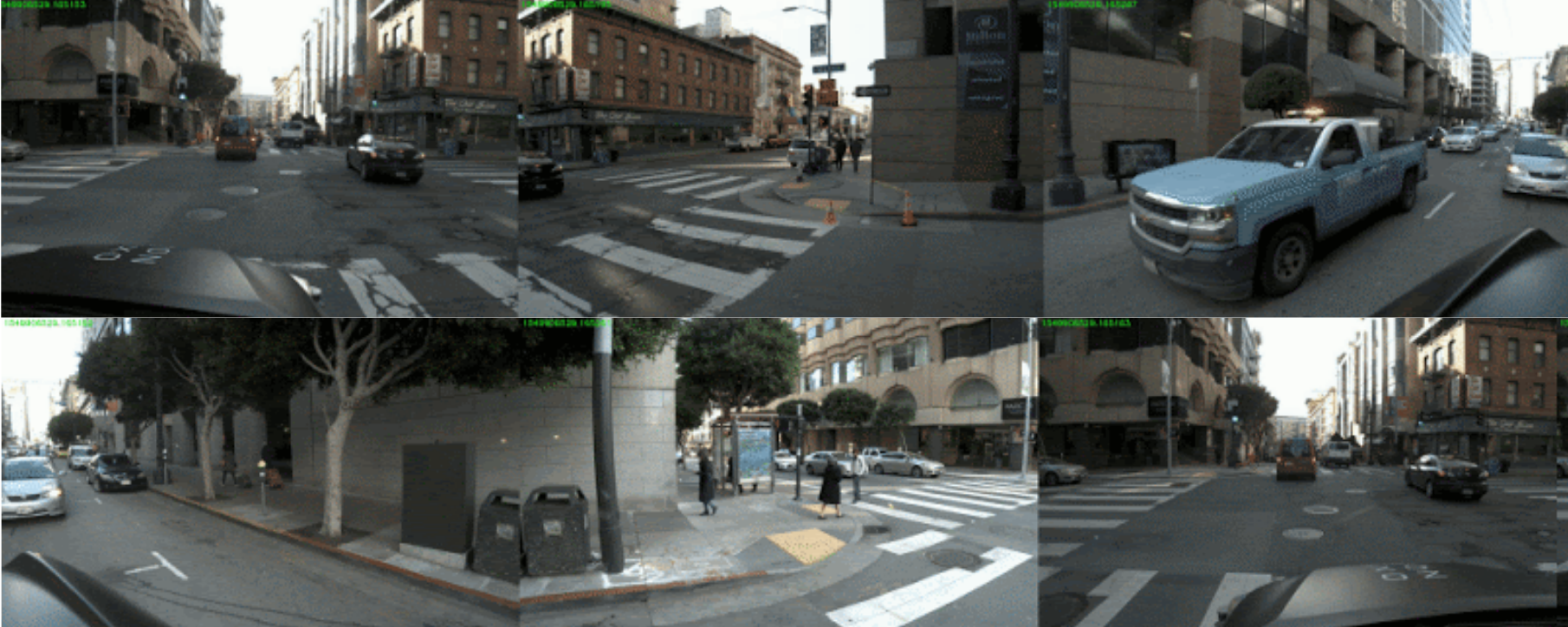}
\end{center}
\caption{Vision Input}
\label{vision_input}
\end{figure}

}
\begin{figure*}[t]
\begin{center}
\includegraphics[width=0.9\linewidth]{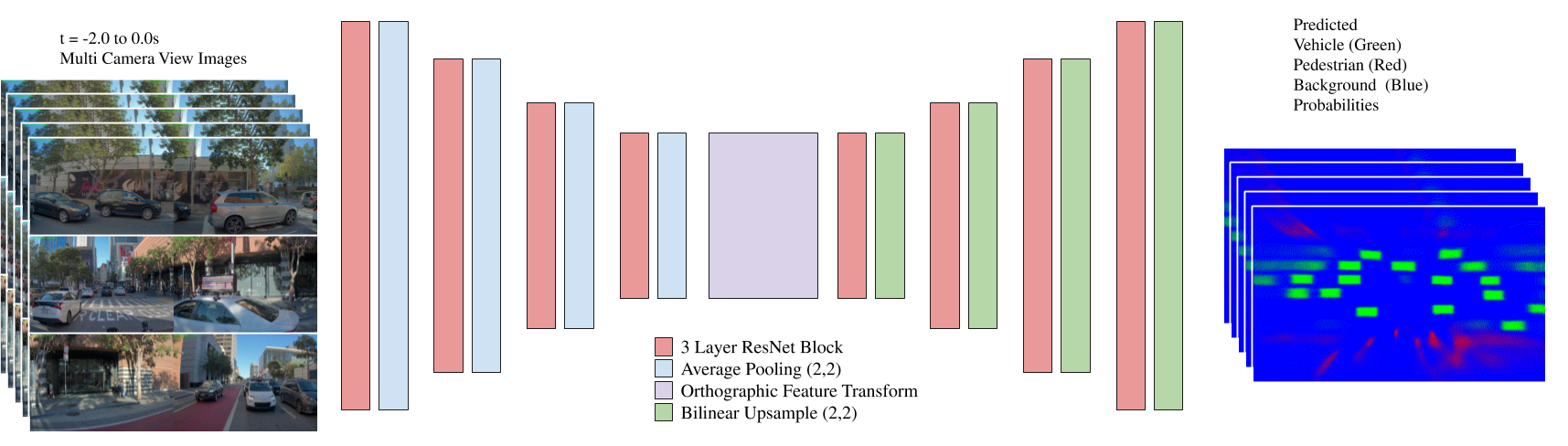}
\includegraphics[width=0.9\linewidth]{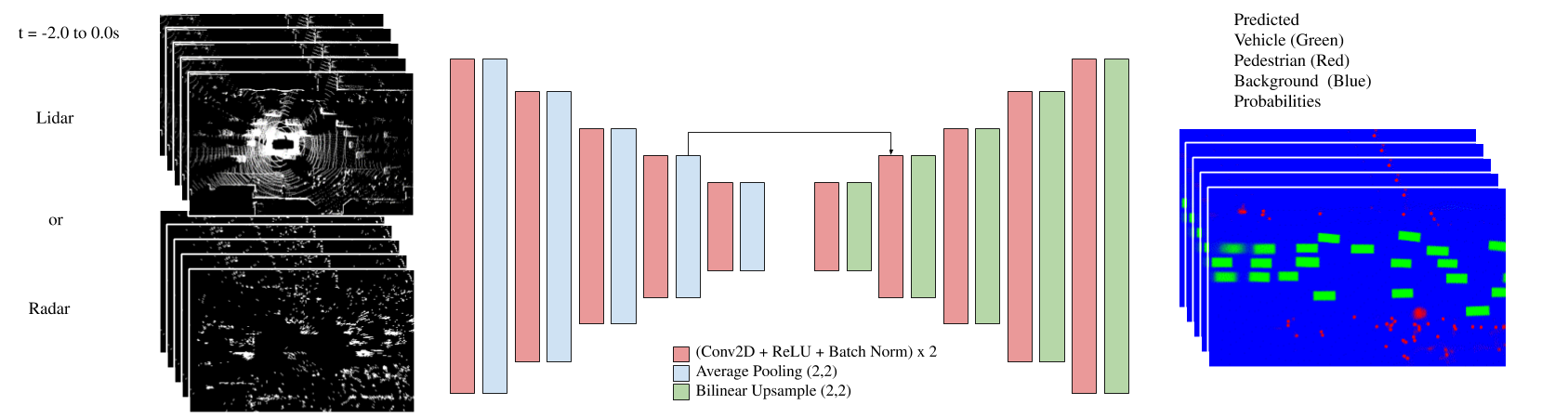}
\end{center}
\caption{Vision architecture on top. Lidar and Radar Architecture on bottom.}
\label{unet}
\end{figure*}

 All of the lidar points for a given sample are transformed into the ego-centric reference frame of the ego vehicle at $t_0$. This transformation is computed based on the ego vehicle position at the corresponding timestep.

\subsection{Radar Input}
Our radar input features are extracted in a similar fashion to the lidar features. The data is captured by radar sensors mounted around the ego vehicle. It consists of 6 top-down grid features.

These features consist of: 1) Binary radar occupancy (1 if any radar point is present a given grid cell, 0 otherwise). 2) X, Y values for each radar return's doppler velocity compensated with ego vehicle's motion. 3)  Radar cross section (RCS). 4) Signal to noise ratio (SNR). 5) Ambiguous Doppler interval.
These features are similarly stacked along the channel dimension.
Radar points for a given sample are also transformed into the ego-centric reference frame at $t_0$.

Figure \ref{radar_input} is a quiver plot of a single frame of the radar Doppler velocity features. A single example consists of 5 of these frames.

\subsection{Vision Input}
Our vision neural network consumes a sequence of images captured by multiple wide angle cameras situated at different angles on the ego vehicle providing a 360\degree field of view.
The dimensions of the images match the output resolution of 192 by 320. Figure \ref{vision_input} is an example of a single timestep of vision views, concatenated side by side.


\subsection{Architecture}

Our overall architecture consists of a neural network for each sensor modality. Across all modalities, our network architecture consists of an encoder decoder network with convolutional layers. We use average pooling with a pooling size of (2,2) in the encoder and up-sampling in the decoder. After the decoder we have a single linear convolutional layer to produce logits, and apply a softmax to produce the final output probabilities for each of the three classes along each of the output timesteps. We use a slightly different encoder and decoder scheme for the vision network compared to the lidar and radar networks to account for the pixel space features. 

\subsection{Lidar/Radar Encoder Decoder}

Lidar and radar rely on a U-Net-like architecture\cite{UNetPaper}. The encoder consists of a set of 5 blocks consisting of a pair of convolutional layers with batch normalization followed by a average pooling layer  (Figure \ref{unet}). The decoder consists of 5 blocks consisting of 3 convolutional layers with batch normalization. The network also includes a skip connection from the fourth block of the encoder to the second block of the decoder in U-Net style.

\subsection{Vision Encoder Decoder}

The vision network architecture consists of a traditional encoder decoder scheme based on a fully convolutional ResNet backbone \cite{resnet} (Figure \ref{unet}). We use a three layer ResNet block with 4 blocks in the encoder and decoder. Between the encoder and decoder we apply an orthographic feature transform layer.

\subsubsection{Pixel Space to Top-Down Transform}

The desired output representation is in top-down view.  However, the inputs for vision are images in perspective view.
To convert from pixel space to top-down space we propose a learned orthographic feature transformation by modifying the View Parsing Network module in \cite{CrossView}.
We implement this transformation as a network layer which operates between the encoder and decoder sections of the network. The orthographic layer consists of a series of unbiased fully connected layers with ReLU activations. This provides the non-linear property of the transformation.  Due to the different reference frame for each camera we pass each camera view through a shared encoder, add together projected features and pass the result through a single decoder.
This method allows us to learn features in pixel space which can then be decoded into a top-down output.

\subsection{Multi-Modal Fusion}
All modalities share the same output reference frame and representation, so we aggregate by applying an aggregation function to the softmax values across the modality dimension. The two aggregation functions we evaluated were average and priority pool.
Average is the normal arithmetic mean, which serves to reduce the variance of the output.
Priority pool is defined as follows: Set priority = {pedestrian: 3, vehicle: 2, background: 1}. If modalities disagree on a pixel classification, we pick the modality whose predicted class has the highest priority value and we break ties by larger magnitude of the softmax value for that object class.

\subsection{Training}
We train each modality independently using an Adam optimizer \cite{kingma2014adam}. The loss function consists of the sum of the cross entropy for all of the output timesteps, with an equal weight for each frame. Additionally, we place an extra factor of 10 on the pedestrian/bike loss to account for class imbalance. 

$$ \mathcal{LOSS} = - \sum_{f=0}^{F} \sum _{g \in {\mathcal {G}}} \sum_{c \in \mathcal{C}} k_c * p_f^g(c)\,\log q_f^g(c) $$

Where $F$ is the number of future frames, $\mathcal{G}$ is the set of cells in the semantic grid, $k_c$ is 10 if $c=\text{ped}$ and 1 otherwise, and $p_f^g(c)$ and $q_f^g(c)$ are the ground truth and predicted values for a given frame $f$, object class $c$ and grid cell $g$.
Additionally, we batch balance based on the yaw of surrounding agents to get a good coverage of various orientations.

\renewcommand{\arraystretch}{1.2}

\begin{table*}[] \scriptsize
\begin{tabular}{|c|l|l|l|l|l|l|l|l|l|l|l|l|l|l|}
\hline
\multicolumn{2}{|l|}{}                                        & \multicolumn{4}{c|}{NuScenes}                                    & \multicolumn{4}{c|}{Lyft}                                        & \multicolumn{5}{c|}{Purpose Built Dataset}                                 \\ \hline
Class                                             & Metric    & Lidar            & Vision  & Average          & Pool             & Lidar            & Vision  & Average          & Pool             & Lidar            & Radar   & Vision  & Average          & Pool             \\ \hline
\multirow{4}{*}{\rotatebox[origin=c]{90}{Pedestrian}}                       & Precision & 70.3\%          & 14.2\% & \textbf{71.3\%} & 40.8\%          & 58.1\%          & 11.0\% & \textbf{70.4\%}  & 30.6\%          & 45.7\%          & 24.6\% & 14.6\% & \textbf{51.3\%} & 20.9\%          \\ \cline{2-15} 
                                                  & Recall    & 38.5\%          & 7.4\%  & 8.5\%           & \textbf{41.1\%}  & 53.8\%          & 11.5\% & 20.3\%          & \textbf{56.9\%} & 84.3\%          & 46.8\% & 28.7\% & 57.0\%          & \textbf{87.1\%} \\ \cline{2-15} 
                                                  & IOU       & \textbf{24.9\%}  & 4.8\%  & 7.6\%            & 20.4\%          & \textbf{38.8\%} & 5.9\%  & 18.7\%          & 24.8\%          & \textbf{42.2\%}  & 19.2\% & 10.7\% & 37.0\%          & 20.3\%          \\ \cline{2-15} 
                                                  & Accuracy  & \textbf{99.8\%} & 99.7\% & 99.8\%          & 99.7\%          & \textbf{99.9\%} & 99.8\% & 99.9\%          & 99.8\%          & 99.8\%          & 99.8\% & 99.7\% & \textbf{99.9\%}  & 99.6\%          \\ \hline
\multirow{4}{*}{\rotatebox[origin=c]{90}{Vehicle}}                          & Precision & 91.2\%          & 69.9\% & \textbf{92.6\%} & 75.8\%           & \textbf{94.7\%} & 78.5\% & 94.6\%           & 81.5\%          & 91.7\%          & 87.0\% & 84.5\% & \textbf{93.6\%} & 78.8\%          \\ \cline{2-15} 
                                                  & Recall    & 86.3\%          & 52.6\% & 72.2\%          & \textbf{88.9\%} & 90.5\%          & 66.1\% & 81.3\%          & \textbf{92.7\%} & 92.1\%          & 80.4\% & 75.6\%  & 87.4\%          & \textbf{95.6\%} \\ \cline{2-15} 
                                                  & IOU       & \textbf{44.3\%} & 30.0\% & 40.6\%           & 40.9\%          & \textbf{86.1\%} & 56.0\%  & 77.7\%          & 76.6\%          & \textbf{85.0\%} & 71.8\% & 66.4\% & 82.5\%          & 76.1\%           \\ \cline{2-15} 
                                                  & Accuracy  & \textbf{99.0\%}  & 96.8\% & 98.4\%          & 98.2\%           & \textbf{99.1\%} & 97.0\% & 98.6\%          & 98.3\%          & \textbf{99.3\%} & 98.6\% & 98.4\% & 99.2\%          & 98.7\%          \\ \hline
\multirow{4}{*}{\rotatebox[origin=c]{90}{Background}} & Precision & 99.2\%          & 97.6\%  & 98.5\%          & \textbf{99.3\%} & 99.4\%           & 97.9\% & 98.8\%          & \textbf{99.5\%} & 99.6\%          & 99.1\%  & 98.8\% & 99.4\%          & \textbf{99.8\%}  \\ \cline{2-15} 
                                                  & Recall    & 99.5\%          & 98.8\% & \textbf{99.7\%} & 98.5\%          & 99.6\%          & 98.8\% & \textbf{99.7\%} & 98.6\%          & 99.5\%          & 99.3\% & 99.2\% & \textbf{99.6\%} & 98.5\%          \\ \cline{2-15} 
                                                  & IOU       & \textbf{49.7\%}  & 49.1\% & 49.5\%          & 49.4\%          & \textbf{99.0\%} & 96.8\%  & 98.5\%          & 98.1\%          & \textbf{99.1\%} & 98.4\% & 98.1\% & 99.1\%           & 98.3\%          \\ \cline{2-15} 
                                                  & Accuracy  & \textbf{98.8\%} & 96.5\% & 98.3\%          & 98.0\%          & \textbf{99.1\%} & 96.9\% & 98.6\%          & 98.2\%          & \textbf{99.2\%} & 98.5\%  & 98.1\% & 99.1\%          & 98.4\%          \\ \hline
\end{tabular}
\vspace*{2mm}
\caption{Table shows 1 vs all Accuracy per class, intersection over union (IOU), precision and recall for all 3 datasets at $t_0$. Purpose built dataset and NuScenes had $k_{c=ped} = 10$ for pedestrian/bike class while Lyft had $k_{c=ped} = 50$.}
\label{results_table}
\end{table*}

\renewcommand{\arraystretch}{1}

\section{Experiments}


We train and evaluate our model on three large-scale datasets. For all datasets we pick the number of past timesteps $p=4$ and the number of future timesteps $f=4$ with an interval of 0.5 seconds in between, using data from timesteps $t=(-2,-1.5,-1,-0.5,0)s$ to infer semantic grids at timesteps $t=(0,0.5,1,1.5,2)s$.


The first two datasets are NuScenes\cite{caesar2019nuscenes} and Lyft \cite{lyft2019}. Both of which are open-source datasets. NuScenes dataset consists of 1000 20 second long scenes captured in Boston and Singapore. Each scene is annotated at 2 Hz thus giving us 40 annotated frames per scene. The dataset contains images captured from 6 cameras, full lidar spin point clouds, 3D annotated bounding boxes, 23 object classes and more. Each frame's ego pose is also recorded allowing us to transform point clouds to different reference frames. We show results of training on ~24,000 samples. Lyft provides sensor data from both lidar and 6 cameras annotated at 10 Hz. We show results of training on  ~22,000 samples. 



The last dataset is purpose-built for this experiment. It consists of sensor data collected with vehicles outfitted with multiple lidars, radars, and cameras with 360\degree coverage. The lidars complete a single spin every 100ms, resulting in a frame rate of 10Hz. Each radar scans the surroundings at 15Hz. The cameras are also sampled at a 10Hz frequency. Our ground truth is generated from a multi-modal perception pipeline which produces 3d track boxes and semantic labels. We then project these boxes into top-down space to generate our output labels.

Our dataset consists of \textasciitilde 2,000,000 samples.
We use a semantic grid of dimensions $192$ in $x$ and $320$ in $y$. For NuScenes and Lyft, we use a 10 $cm$ resolution for each grid cell giving a coverage of 614 $m^2$.  The purpose-built dataset uses a 20 $cm$ resolution for each grid cell giving a coverage of 2457 $m^2$.

\section{Results}
We evaluate our models using a 10\% validation split, and generate metrics on a per grid-cell and per class basis.

\subsection{NuScenes Dataset}
We evaluate using lidar and vision networks on NuScenes after we've found that its radar data were too sparse to perform semantic segmentation. Lidar achieves highest performance on precision and recall for all three classes, which is expected since all of the lidar input features are in top-down and the lidar point clouds are represented with respect to the ego-position at $t_0$. The camera network, on the other hand, has a more challenging task on two fronts. One, the camera model has to learn the mapping from pixel-space to a top-down space while lacking any depth information. Secondly it has to learn the ego motion model. We aggregate the resulting prediction using average and priority pooling. Average pooling achieves higher precision while priority pooling achieves higher recall (except for the lowest priority class which is background).
Figure \ref{nuscenes_plots} shows the precision and recall per prediction horizon for each class. In all classes, performance declines as prediction horizon increases. Pedestrians prove to be the hardest to learn and predict into the future.  

\subsection{Lyft Dataset}
One difference observed in Lyft dataset is that there are much fewer pedestrian samples (including bicycles and motorcycles) compared to NuScenes dataset. Therefore a larger class weight of 50 is assigned to the pedestrian class during training. 

The performance of both modalities is evaluated in Table \ref{results_table}. We are able to achieve comparable performance in Lyft dataset as in NuScenes dataset. The Lidar model performs better than the vision model in both precision and recall for all three classes. The aggregated results using average and priority pooling are demonstrated as well.

\subsection{Purpose-Built Dataset}

We also evaluate our model using purpose built datasets of varying sizes. Naturally, we found that the larger dataset led to significant performance improvements across all object classes and time frames.


Comparing the three modalities, lidar achieves the best performance, with radar a distant second and vision close behind radar. These results are highlighted in Table \ref{results_table}. These results can be compared to the NuScenes results in Table \ref{results_table}, where the purpose-build dataset results are better across the board.
Similar to NuScenes and Lyft results, average pooling achieves the highest precision and priority pool achieves the highest recall.

Figure \ref{plots} shows accuracy, intersection over union (IOU), precision and recall plotted against prediction horizon for vehicle and pedestrian object classes. All modalities show a monotonic decrease as the prediction horizon increases. Pedestrian precision notably falls a lot faster as the prediction horizon increases compared to vehicle precision. This indicates the difficulty of predicting pedestrian motion compared to vehicles.

\begin{figure*}[t]
\begin{center}
  \includegraphics[width=0.9\linewidth]{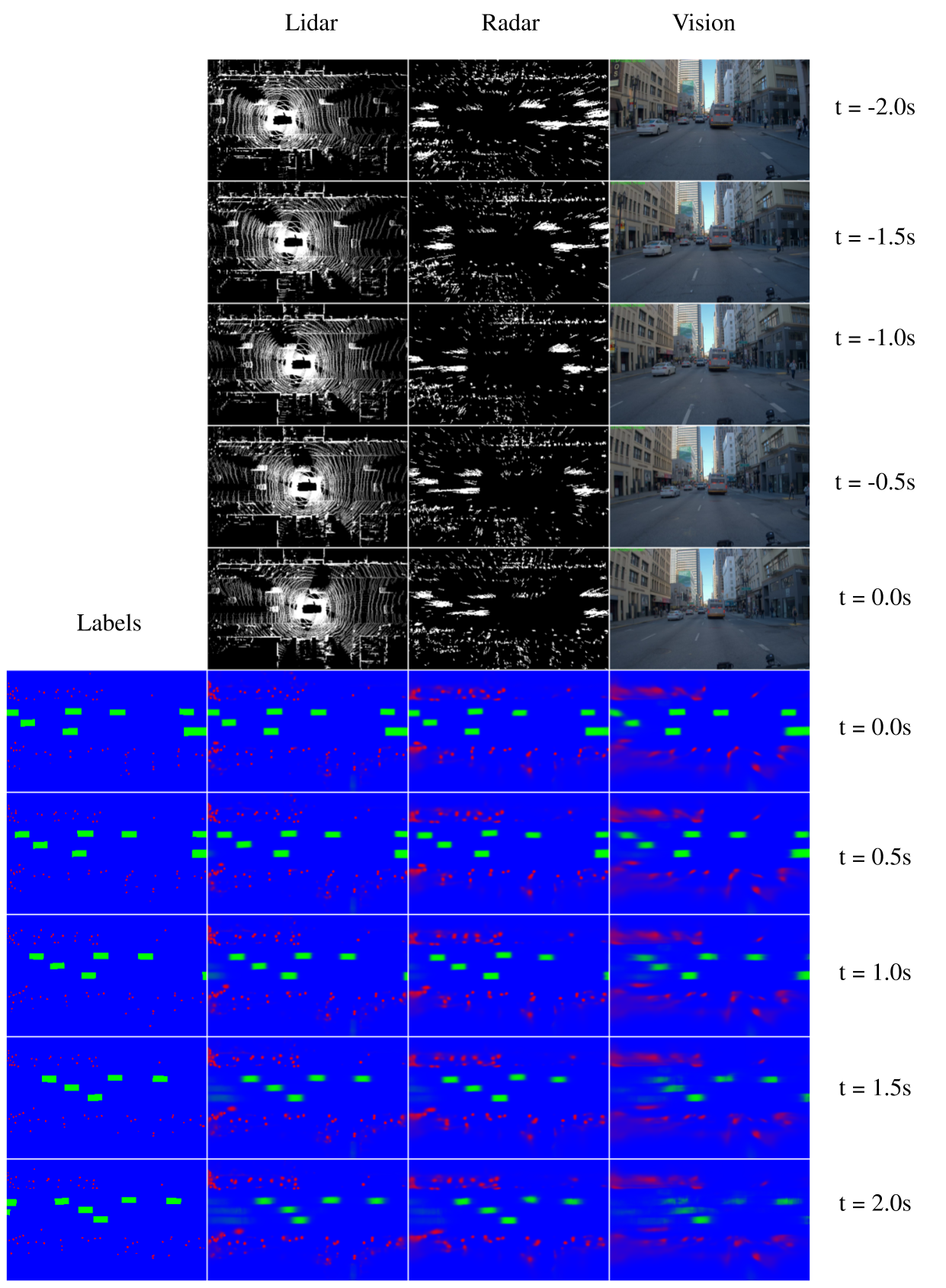}
\end{center}
\caption{Label (lower left), input (top) and predictions (bottom) for lidar radar and vision. One of the six input cameras is visualized.}
\label{stacked}
\end{figure*}

Figure \ref{stacked} shows labels, input and output at all frames for all three modalities. Lidar has the highest resolution for fine detail such as pedestrians. Vision and radar tend to lose the details when many agents are stacked next to each other. Radar does a good job of depth when compared to vision. Radar also handles kinematics relatively well which may be due in part to the velocity measurements. Vision tends to struggle with predicting depth properly, and predictions are often smeared across the radial dimension.



\section{Conclusion and Future Work}
We present a top-down framework that takes in sensor data from different modalities and performs semantic segmentation for both the current timestep and the near future. By using different sensor modalities such as lidars, radars, and cameras from NuScenes and our purpose-built datasets, this work demonstrates the framework's ability to adapt to different sensor configurations. A common representation across modalities simplifies the complexity of the perception system and provides flexibility in choosing fusion strategies based on different performance requirements. On top of that, by using the same framework, the semantic segmentation tasks can be expanded further from current timestep into future timesteps, allowing a short horizon prediction of road users' motions. The extensible nature of this framework provides flexibility and robustness for downstream decision making and motion planning.

One limitation of our approach is that we use a fully grid based representation with no object level representation.
We hope to extend this work to add instance segmentation and box regression in the top-down representation.

{
\begin{figure*}[!htbp]
\begin{center}
\subfloat{\includegraphics[width = 0.35\linewidth]{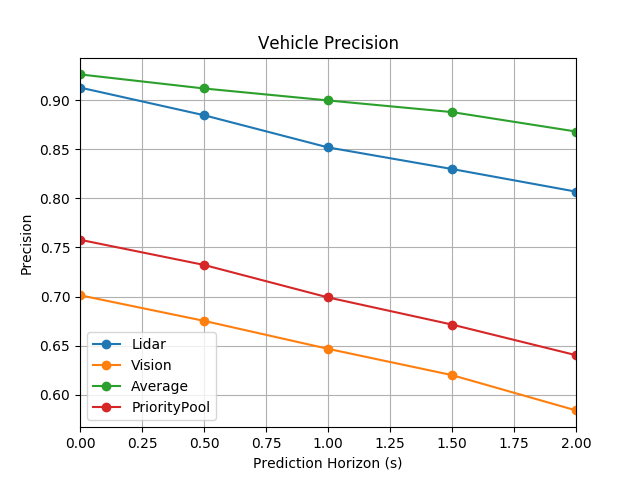}}
    \hspace{0.1\linewidth}
\subfloat{\includegraphics[width = 0.35\linewidth]{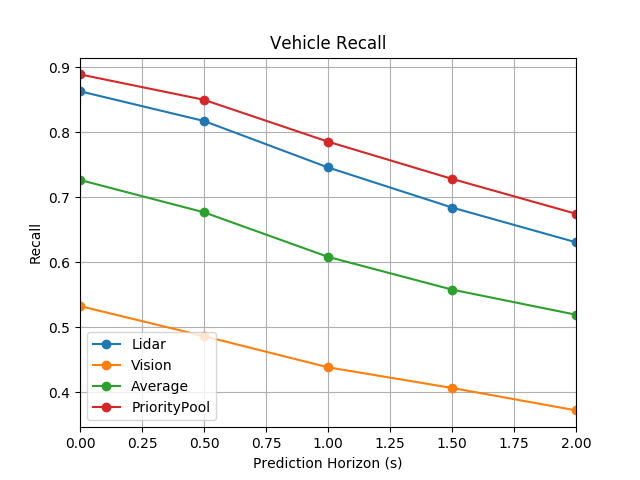}}\\
\subfloat{\includegraphics[width = 0.35\linewidth]{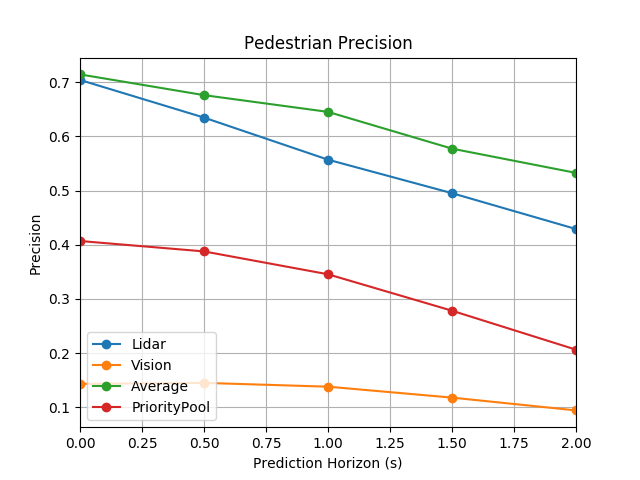}}
    \hspace{0.1\linewidth}
\subfloat{\includegraphics[width = 0.35\linewidth]{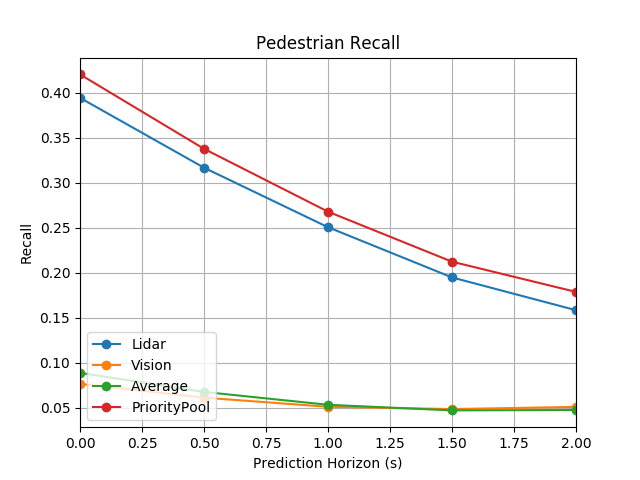}}
\caption{Precision and Recall Metrics for NuScenes dataset}
\label{nuscenes_plots}
\end{center}
\end{figure*}
}
{
\begin{figure*}[t]
\begin{center}
\subfloat{\includegraphics[width = 0.35\linewidth]{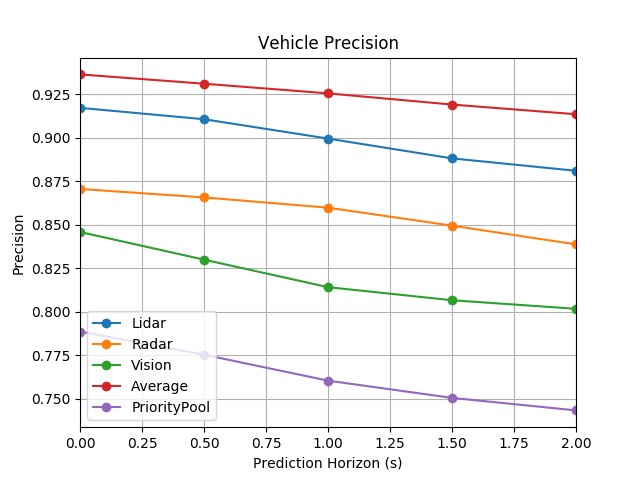}}
    \hspace{0.1\linewidth}
\subfloat{\includegraphics[width = 0.35\linewidth]{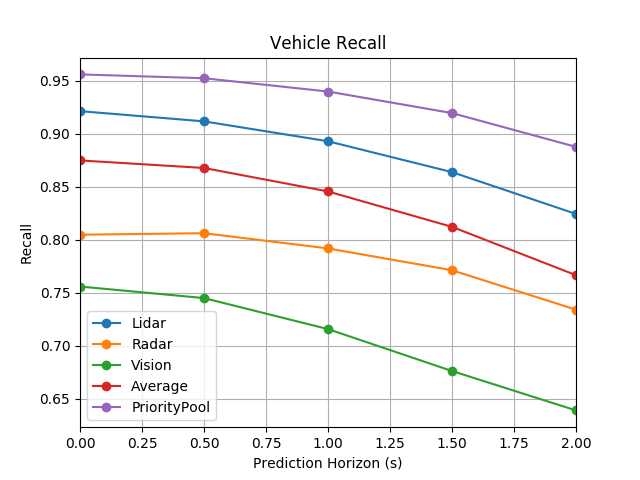}}\\
\subfloat{\includegraphics[width = 0.35\linewidth]{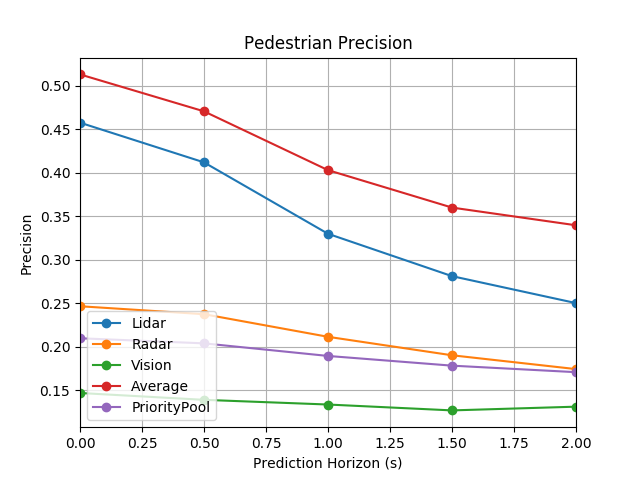}}
    \hspace{0.1\linewidth}
\subfloat{\includegraphics[width = 0.35\linewidth]{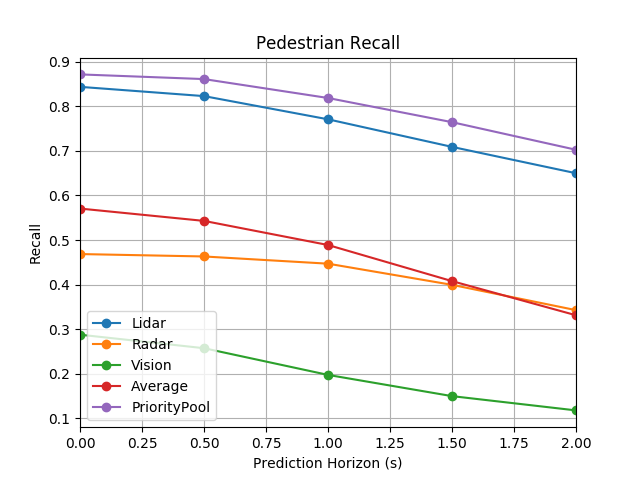}}
\caption{Precision and Recall Metrics for Purpose-Built dataset}
\label{plots}
\end{center}
\end{figure*}
}

\newpage
\newpage
{\small
\bibliographystyle{ieee_fullname}
\bibliography{egbib}
}

\end{document}